%% file: acl_latex.tex
\pdfoutput=1
\documentclass[11pt]{article}

\usepackage[final]{acl}

\usepackage{hyperref}

\usepackage{float}    % For H placement specifier

\usepackage[ruled,vlined]{algorithm2e}
\usepackage[compatible]{algpseudocode} % or \usepackage{algcompatible}

\usepackage{wrapfig}

\usepackage[ruled,vlined]{algorithm2e}
\usepackage{bm}
\usepackage{xcolor}
\SetCommentSty{mycommfont}
\SetKwInput{KwInput}{Input}
\SetKwInput{KwOutput}{Output}

\usepackage{times}
\usepackage{textcomp}
\usepackage{xcolor}
\usepackage{latexsym}
\usepackage{multirow}
\usepackage{hyperref}
\usepackage{amsmath}
\usepackage{amssymb}
\usepackage{amsthm}
\usepackage{subcaption}

\usepackage{bm}
\usepackage{xcolor}
\usepackage{pifont}
\usepackage{amsmath}
\usepackage{amssymb}
\usepackage{svg}

\newcommand{\ours}{OWS }
\usepackage{booktabs} % For formal tables
\usepackage{multirow} % For combining rows
\usepackage{subcaption}
\usepackage{tikz}
\newcommand*\circled[1]{\tikz[baseline=(char.base)]{
            \node[shape=circle,draw,inner sep=0.4pt] (char) {#1};}}
\usepackage{enumitem}
\usepackage[utf8]{inputenc} % allow utf-8 input
\usepackage[T1]{fontenc}    % use 8-bit T1 fonts
\usepackage{hyperref}       % hyperlinks
\usepackage{url}            % simple URL typesetting
\usepackage{booktabs}       % professional-quality tables
\usepackage{amsfonts}       % blackboard math symbols
\usepackage{nicefrac}       % compact symbols for 1/2, etc.
\usepackage{microtype}      % microtypography
\usepackage{xcolor}         % colors
\newcommand{\mbf}{\mathbf}%
\usepackage{colortbl}
\newcommand{\gr}{\rowcolor[gray]{.95}}

\definecolor{commentcolor}{RGB}{110,154,155}   %

\newlength\myindent
\usepackage[textsize=tiny]{todonotes}

\title{Outlier-weighed Layerwise Sampling for LLM Fine-tuning}

\author{
Pengxiang Li$^{1,*}$ \quad
Lu Yin$^{2,3,*}$ \quad
Xiaowei Gao$^{4}$ \quad
Shiwei Liu$^{5,3,\dagger}$ \\
$^{1}$Dalian University of Technology \quad
$^{2}$University of Surrey \quad
$^{3}$Eindhoven University of Technology \\
$^{4}$University College London \quad
$^{5}$University of Oxford \\
\texttt{shiwei.liu@maths.ox.ac.uk}%
\thanks{Equal contribution.\quad $^{\dagger}$Corresponding author.}
}
\begin{document}
\maketitle
\begin{abstract}
The rapid advancements in Large Language Models (LLMs) have revolutionized various natural language processing tasks. However, the substantial size of LLMs presents significant challenges in training or fine-tuning. While parameter-efficient approaches such as low-rank adaptation (LoRA) have gained popularity, they often compromise performance compared to full-rank fine-tuning. In this paper, we propose \textit{Outlier-weighed Layerwise Sampling} \textbf{\textit{(OWS)}}, a new memory-efficient fine-tuning approach, inspired by the layerwise outlier distribution of LLMs.
Unlike LoRA, which adds extra adapters to all layers, OWS strategically assigns higher sampling probabilities to layers with more outliers, selectively sampling only a few layers and fine-tuning their pre-trained weights. To further increase the number of fine-tuned layers without a proportional rise in memory costs, we incorporate gradient low-rank projection, further boosting the approach’s performance. Our extensive experiments across various architectures, including LLaMa2, and Mistral, demonstrate that OWS consistently outperforms baseline approaches, including full fine-tuning. Specifically, it achieves up to a 1.1\% average accuracy gain on the Commonsense Reasoning benchmark, a 3.0\% improvement on MMLU, and a notable 10\% boost on MT-Bench, while being more memory efficient. OWS allows us to fine-tune 7B LLMs with only 21GB of memory. Our code is available at \url{https://github.com/pixeli99/OWS}. 
\end{abstract}

\input{secs/1_intro}

\input{secs/2_related}

\input{secs/3_OWS}
\input{secs/4_exps}

\section{Limitations}
\label{sec:limitations}
While our proposed \textbf{OWS} approach demonstrates notable improvements in fine-tuning performance and memory efficiency, there are some factors to consider. Our evaluation primarily focuses on LLaMa2 and Mistral models, and further research could investigate the generalizability of \textbf{OWS} across a broader range of models and architectures.

% As with any fine-tuning technique, the application of \textbf{OWS} in real-world settings may involve certain considerations. In particular, selective layer updates might introduce some biases in model behavior or lead to slight variations in performance for tasks that were less represented during training. These aspects highlight the need for further exploration and testing in diverse settings to ensure consistent and reliable performance.

\section*{Acknowledgements}
S. Liu is funded by the Royal Society with the
Newton International Fellowship.

\bibliography{main}

\clearpage
\appendix
\onecolumn

\section{OWS-Reverse}
\label{app:reverse}

To further validate our approach, we introduce a new baseline: OWS-Reverse. This variant assigns lower sampling probabilities to layers with a higher proportion of outliers. As expected, OWS-Reverse performs the worst among the tested fine-tuning strategies, reinforcing our intuition about the importance of outlier-weighted prioritization in achieving better results.

\begin{table}[h]
\centering
\caption{Comparison with varies baselines.}
\resizebox{0.9\textwidth}{!}{%
\begin{tabular}{@{}lcccccccccc@{}}
\toprule
\textbf{Method} & \textbf{MMLU} & \textbf{BoolQ} & \textbf{PIQA} & \textbf{SIQA} & \textbf{HellaSwag} & \textbf{WinoGrande} & \textbf{ARC-e} & \textbf{ARC-c} & \textbf{OBQA} & \textbf{Avg.} \\
\midrule
OWS-Reverse & 49.4 & 81.9 & 77.8 & 33.4 & 59.1 & 80.1 & 79.3 & 50.2 & 38.2 & 61.0 \\
Galore & 49.6 & 81.8 & 79.4 & 32.9 & 60.7 & 79.6 & 79.8 & 49.4 & 37.6 & 61.2 \\
LISA & 49.6 & 82.0 & 79.9 & 33.5 & 59.7 & 79.6 & 80.4 & 51.1 & 38.8 & 61.6 \\
OWS & 52.6 & 85.4 & 80.7 & 34.2 & 60.3 & 82.2 & 80.6 & 51.0 & 39.1 & 62.9 \\
\bottomrule
\end{tabular}%
}
\label{tab:baseline_comparison}
\end{table}

\section{Hyperparameter Analysis}

\begin{figure}[h]
    \centering
    \includegraphics[width=0.6\linewidth]{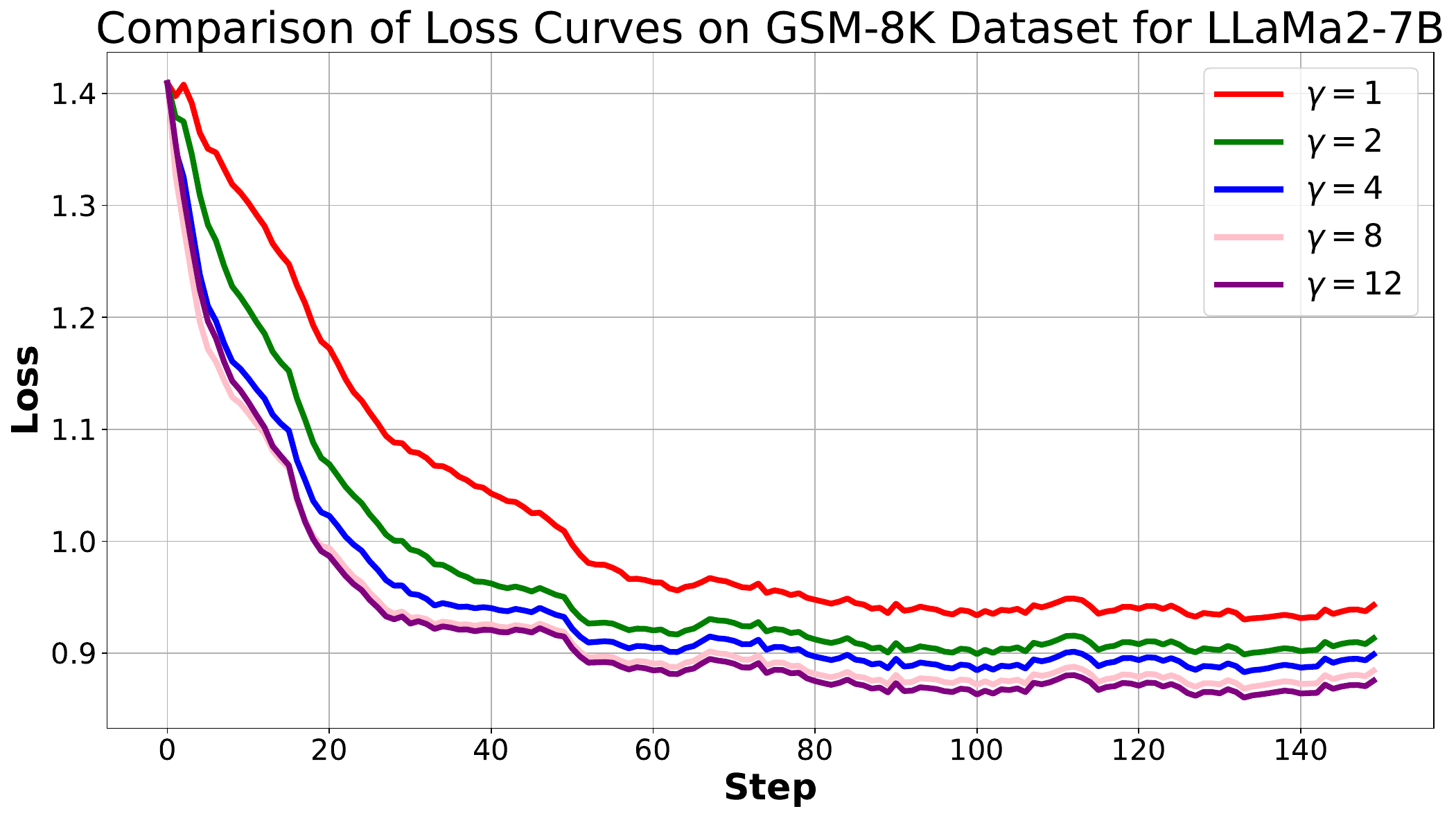}
        % \vspace{-1em}
    \caption{{Fine-tuning loss of LLaMA2-7B using method OWS on the GSM-8K dataset with various sampled layers.}}
    \label{fig:sample_layers}
\end{figure}

$\tau$ is the key hyperparameter to obtain the outlier ratio and sampling layers $\gamma$ is also crucial to OWS  To obtain intuitive and empirical guidance on these hyperparameter choices, we conduct ablation studies using LLaMA2-7B models with the GSM-8K dataset and report the results below.

\begin{table}[ht]
\centering
\caption{GSM scores for different $\tau$ values}
\resizebox{\textwidth}{!}{% Resize table to fit within the text width
\setlength{\tabcolsep}{5mm} % Adjust the column spacing
\begin{tabular}{@{}lccccccccc@{}}
\toprule
Setting & $\tau=3$ & $\tau=5$ & $\tau=7$ & $\tau=9$ & $\tau=11$ & $\tau=13$ & $\tau=15$ & $\tau=17$ & $\tau=19$ \\
\midrule
GSM Scores & 19.18 & 19.41 & 20.04 & 20.62 & 21.15 & 20.24 & 20.17 & 20.47 & 19.79 \\
\bottomrule
\end{tabular}
}
\label{GSM_tau}
\end{table}

We found that mid-range values of $\tau$, such as 9, 11 and 13, generally lead to better performance. This may stem from the fact that the outliers screened by these values are more indicative of heavy-tailed properties. By default, we choose $\tau=13$ for all experiments of OWS.

As for the sampling layer $\gamma$, it is not surprising that performance improves consistently with the sampling of more layers. OWS outperforms LISA with less memory usage across all sampling layer counts. This is attributed to OWS's allocation of higher sampling probabilities to layers abundant in outliers, combined with its efficient low-rank gradient updating technique.

The training curve across different values of $\gamma$ is depicted in Figure~\ref{fig:sample_layers}. Notably, fine-tuning with a higher $\gamma$ leads to faster convergence and lower loss.
% \begin{figure}[t]
%     \centering
%     \includegraphics[width=0.6\linewidth]{figs/loss_sample_layer.pdf}
%         % \vspace{-1em}
%     \caption{\textcolor{blue}{Fine-tuning loss of LLaMA2-7B using method OWS on the GSM-8K dataset with various sampled layers.}}
%     \label{fig:sample_layers}
% \end{figure}

\section{Statistical Significance Test}

We conducted experiments using 5 different seeds and reported the corresponding standard deviations. We do experiments with LISA and OWS to demonstrate the effectiveness of our proposed approach. For MT-Bench, we provided the results evaluated using GPT-4o.

\begin{table*}[htbp]
\centering
\caption{Results of experiments for different models and methods evaluated on MT-Bench with 5 seeds and reported standard deviations.}
\vspace{-1em}
\resizebox{0.9\textwidth}{!}{%
\begin{tabular}{@{}lccccccccc@{}}
\toprule
\textbf{Model} & \textbf{Method} & \textbf{BoolQ} & \textbf{PIQA} & \textbf{SIQA} & \textbf{HellaSwag} & \textbf{WinoGrande} & \textbf{ARC-e} & \textbf{ARC-c} & \textbf{OBQA} \\
\midrule
LLaMa2-7B & LISA & 81.9 ± 0.22 & 79.6 ± 0.26 & 33.6 ± 0.11 & 59.6 ± 0.09 & 79.5 ± 0.16  & 80.3 ± 0.13 & 51.1 ± 0.11 & 39.2 ± 0.12 \\
LLaMa2-7B & OWS  & 85.3 ± 0.19 & 80.8 ± 0.29 & 34.2 ± 0.14 & 60.2 ± 0.11 & 82.4 ± 0.18  & 80.8 ± 0.14 & 51.1 ± 0.12 & 39.6 ± 0.15 \\
Mistral-7B & LISA & 84.9 ± 0.21 & 82.7 ± 0.21 & 33.4 ± 0.11 & 64.4 ± 0.14 & 85.7 ± 0.16  & 83.6 ± 0.11 & 54.3 ± 0.10 & 40.5 ± 0.14 \\
Mistral-7B & OWS  & 88.0 ± 0.24 & 84.0 ± 0.23 & 33.9 ± 0.11 & 66.4 ± 0.16 & 85.6 ± 0.09  & 84.1 ± 0.15 & 57.8 ± 0.14 & 40.5 ± 0.13 \\
\bottomrule
\end{tabular}%
}
\vspace{-0.2em}
\label{tab:model_results}
\end{table*}

Additionally, we performed an independent samples t-test to assess the statistical significance of the performance difference between OWS and LISA. For example, in the LLaMa2-7B model, the t-test yields a t-statistic of -11.36 and a p-value of 3.41e-06, indicating that the performance improvements of OWS over LISA are statistically significant.

\begin{table}[h]
\centering
\caption{Independent samples t-test results for the performance differences between OWS and LISA.}
\resizebox{0.3\textwidth}{!}{%
\begin{tabular}{@{}lcc@{}}
\toprule
\textbf{Model} & \textbf{t-statistic} & \textbf{p-value} \\
\midrule
LLaMa2-7B  & -11.36 & 3.41e-06 \\
Mistral-7B & -13.46 & 9.32e-07 \\
\bottomrule
\end{tabular}%
}
\label{tab:t_test_results}
\end{table}

\clearpage
\newpage

\section{Training Configurations of OWS}
\label{app:hyper}

We utilize Hugging Face and PyTorch for the implementation of our work.
\begin{table}[h]
    \centering
    \caption{Hyperparamters used of OWS for fine-tuning LLaMa2-7B and Mistral-7B on the Commonsense Reasoning Benchmark.}
    \resizebox{0.5\linewidth}{!}{
    \begin{tabular}{c|c|c}
    \toprule
        \textbf{Hyperparameter} & \textbf{LLaMa2-7B} & \textbf{Mistral-7B} \\
    \midrule
        Batch Size             & 16        & 16         \\
        Max. Sequence Length   & 512       & 512        \\
        Learning Rate          & 3e-4      & 3e-5       \\
        Schedular              & linear    & linear     \\
        Training Epoch         & 1         & 1          \\
        Warmup Steps           & 0         & 0          \\
        dtype                  & bfloat16  & bfloat16   \\
    \bottomrule
    \end{tabular}}
    \label{tab:hyperparameters_ft}
\end{table}

\begin{table}[htbp]
    \centering
    \caption{Hyperparamters used of OWS for fine-tuning LLaMa2-7B on various benchmarks.}
    \resizebox{0.8\linewidth}{!}{%
    \begin{tabular}{c|cccc}
        \toprule
        \textbf{Benchmarks} &  \textbf{Commonsense Reasoning} & \textbf{MT-Bench} & \textbf{MMLU} & \textbf{GSM8K}\\
        \midrule
        Train Samples    & 170K & 52K  & 99.8K & 7.4K \\
        Test Samples     & 22.4K & Alpaca-GPT4 (3.3K) & 14K & 1.3K \\
        Batch Size       & 16 & 16 & 16 & 16 \\
        Max\_length      & 512 & 512 & 512 & 512 \\
        Training Epoch   & 1   & 1   & 1   & 1 \\
        Learning Rate    & 3e-4 & 3e-4 & 3e-4 & 3e-4\\
    \bottomrule
    \end{tabular}}
    \label{tab:hyperparameters_downstream}
\end{table}

\section{Pseudocode of GaLore} 
\label{appendix:galore}

Following we present the pseudocode of OWS.

\begin{algorithm*}[h]
\SetKwInput{KwRequire}{Require}
\small
\caption{Outlier-Weighed Layerwise Sampling (OWS)}
\label{alg:lisa}
\DontPrintSemicolon
\KwRequire{number of layers $N_L$, number of training iterations $T$, sampling period $K$, sampled layers $\gamma$, rank level $r$, and $\mathcal{U}(0,1)$  
 refers to a uniform sampling.}

\textcolor{gray}{\% Before Training\\}
{\For{$\ell \gets 1$ \KwTo $N_L$}{
Calculate outlier ratio $D_j$ using the Equation \ref{eq:lod} \;
$p_\ell \gets \frac{\gamma D_\ell}{\sum_{j=1}^{N_L} D_j}$ \Comment{ \textcolor{blue}{Mapping layerwise outlier distribution to sampling probability.}}
}}

\textcolor{gray}{\% Training\\}
\For{$i \gets 0$ \KwTo $T/K - 1$}{
\For{$\ell \gets 1$ \KwTo $N_L$}{

\If{~$\mathcal{U}(0,1) > p_\ell$}{
Freeze layer $\ell$ \;
}

\Else  {    Update the weights in layer $\ell$ \Comment{ \textcolor{blue}{OWS updates the in the low-rank subspace }  }

            {grad = weight.grad}  \\
            {lowrank\_grad = \textbf{project}(grad)} \Comment{\textcolor{blue}{original space -> low-rank space}}  \\
            {lowrank\_update = \textbf{Adam\_update} (lowrank\_grad)}              \Comment{\textcolor{blue}{update by Adam, Adafactor, etc.}}\\
            {update = \textbf{project\_back}(lowrank\_update)}             
            \Comment{\textcolor{blue}{low-rank space -> original space}}   \\
            {weight.data += update} \\
         }
}

}
% \hrulefill

\vspace{-0.5em}
\label{alg:OWS}
\end{algorithm*}
% \vspace{-1em}

\end{document}

%% file: secs/1_intro.tex
\section{Introduction}

The rapid advancements in AI driven by Large Language Models (LLMs) have fundamentally transformed how people work and communicate. The impressive language capabilities of LLMs enable a single model to handle various tasks simultaneously, including but not limited to natural language understanding~\citep{brown2020language,touvron2023llama}, text generation~\citep{kocon2023chatgpt,anil2023palm}, machine translation~\citep{jiao2023chatgpt}, and programming~\citep{surameery2023use,tian2023chatgpt}. However, the massive size of LLMs presents significant challenges for practical applications and deployment.

To address these challenges, various parameter-efficient fine-tuning (PEFT) approaches have been proposed, including prompt tuning~\citep{lester2021power,liu2021gpt}, adaptors~\citep{houlsby2019parameter,he2021towards}, and low-rank adaptation (LoRA) \citep{hu2021lora,dettmers2024qlora}. These approaches enable the fine-tuning of pre-trained LLMs with substantially fewer trainable parameters, making LLM fine-tuning more feasible in practice. Among these, LoRA \citep{hu2021lora} stands out for its re-parameterization technique of the pre-trained weight matrix $W \in \mathbb{R}^{m \times n}$, expressed as $ W_0 + AB$, where $A \in \mathbb{R}^{m \times r}$, $B \in \mathbb{R}^{r \times n}$, and $r \ll \min (m,n)$. By fine-tuning only the low-rank adaptor $AB$ while keeping the pre-trained weight $W_0$ frozen, LoRA significantly reduces the memory usage and computational costs associated with fine-tuning LLMs, rapidly becoming the preferred method for such tasks. Despite its efficiency, recent research has highlighted the inferior performance of low-rank reparameterization compared to full-rank updates in both fine-tuning scenarios~\citep{xia2024chain,biderman2024lora} and pre-training contexts~\citep{lialin2023stack,zhao2024galore}. These findings underscore the need for further exploration into balancing training efficiency with model performance, particularly in the context of large-scale language models.

\begin{figure*}[t]
    \centering
    \hspace{-1em}
    \includegraphics[width=0.99\linewidth]{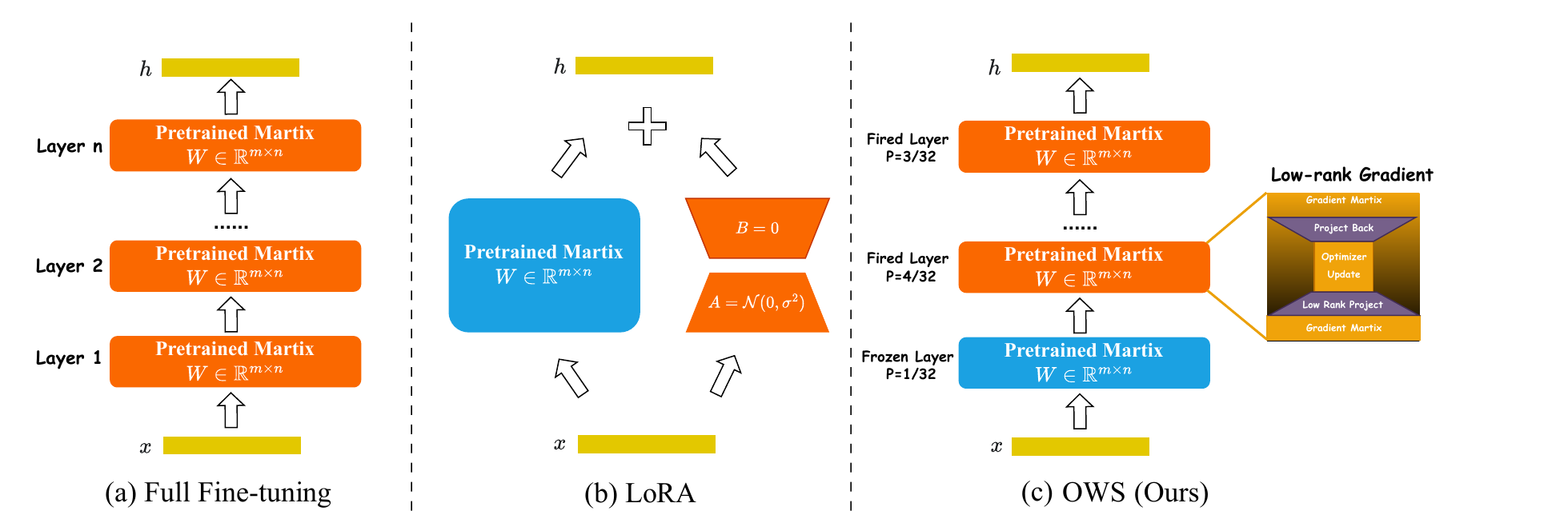}

    \caption{The comparison among Full Fine-tuning, training with LoRA, and OWS. Blue modules are frozen, while orange modules are activated. OWS non-uniformly samples layers to fine-tune models with low-rank gradients.}
    \label{fig:Enhancer}
    \vspace{-1.5em}
\end{figure*}

Recently, Layerwise Importance Sampled AdamW (LISA) \citep{pan2024lisa} has emerged as a promising alternative for LLM fine-tuning, integrating the concept of importance sampling \citep{kloek1978bayesian, zhao2015stochastic} into the fine-tuning process. Unlike methods that add adapters to all layers, LISA selectively samples a small subset of layers and directly fine-tunes their pre-trained weights, demonstrating notable performance improvements over LoRA.
However,  our investigation reveals two limitations of LISA: 
\begin{itemize}[leftmargin=*]
    \item [$\star$] LISA employs random sampling of layers for fine-tuning, which results in suboptimal performance due to the varying importance of layers in LLMs. To illustrate this, we demonstrate that random sampling underperforms compared to a simple baseline—monotonic decreasing sampling probabilities from top to bottom layers—as shown in Table \ref{tab:linear_decreasing}.
    
    \item [$\star$] The fine-tuning of sampled layers is conducted in a full-rank manner, leading to substantial memory overhead as the number of sampled layers increases. While fine-tuning accuracy improves with the inclusion of more sampled layers, this improvement comes at the cost of escalating memory usage, as detailed in Table \ref{tab:various_gamma}.
\end{itemize}

\textbf{Overview.} To address these limitations,  we introduce Outlier-weighted Layerwise Sampling (\textbf{OWS}), a novel layerwise sampling approach to fine-tune LLMs. \ours leverages the unique characteristic of LLMs where certain features and weights—referred to as outliers—have significantly larger magnitudes than the rest \citep{kovaleva2021bert,puccetti2022outliers,dettmers2022llm,yin2023outlier,lu2024alphapruning}. Our rationale is that layers with more outliers play a more active role in learning the data distribution of the text corpus, as they have received larger gradients during pre-training. Therefore, we assign higher sampling probabilities to layers with a greater concentration of outliers, in a way that the features in these layers will be leveraged more frequently to adapt to downstream datasets. 
To further reduce the memory costs of fine-tuning, we update the optimization status of sampled layers in a low-rank subspace, which allows us to increase the number of fine-tuned layers without a proportional rise in memory costs. 

Built upon these techniques, \ours achieves a large performance boost to LISA. Our extensive experiments across commonly used architectures including LLaMa2 \citep{touvron2023llama} and Mistral \citep{jiang2023mistral} demonstrate that OWS consistently outperforms its baseline approaches including full-parameter fine-tuning. OWS achieves up to a 1.1\% average accuracy gain on the Commonsense Reasoning benchmark, a 3.0\% improvement on MMLU, and a notable 10\% boost on MT-Bench.

%% file: secs/2_related.tex
% BAdam utilizes a block coordinate descent (BCD) framework where the entire set of model parameters is divided into blocks. Each block is optimized one at a time using Adam, reducing the memory requirements for each update step. This method is designed for full parameter optimization but achieves memory efficiency by sequentially updating smaller subsets of parameters.

% \textbf{Memory-Efficient Full Parameter Fine-Tuning.} Despite the success of PEFT methods, finetuning within a substantially lower-dimensional subspace may potentially limit downstream performance. Recently, techniques like BAdam \citep{luo2024badam} have been proposed to reduce memory consumption without sacrificing the benefits of full-parameter optimization. BAdam utilizes block coordinate descent with Adam as an inner solver, demonstrating efficiency in both memory use and convergence behavior, making it a viable alternative to LoRA in resource-constrained environments.

%% file: secs/3_OWS.tex
% \vspace{-0.5em}
\section{Background and Motivation}
In this section, we first introduce LISA's algorithm and then present our findings of two key limitations of LISA: the shortcomings of its sampling approach and the significant memory overhead associated with the sampled layers.

\subsection{Background: LISA} 

\citet{pan2024lisa} conducted an in-depth analysis of LoRA's training dynamics across layers and revealed an unusual skew in the distribution of layerwise weight norms, particularly towards the top layer and/or the bottom layer \footnote{Please note that in LISA, the terms 'top' and 'bottom' layers refer to the embedding layer and the LLM head layer, respectively, rather than the first and last Transformer blocks.}, where the norms are significantly larger compared to other layers. Building upon this insight, the authors proposed LISA, a novel fine-tuning approach for LLMs, which incorporates the concept of importance sampling \citep{kloek1978bayesian,zhao2015stochastic} into the fine-tuning process. In LISA, layers of the base model are sampled to be unfrozen during training based on a prescribed probability, with the exception of the top and bottom layers, which remain activated throughout the process. Given a network with $N_L$ layers, the sampling probability of layer $\ell$ is given as follows:
\begin{equation}
p_{\ell}=\left\{
\begin{array}{lcl}
1.0,  &   & {if \; \ell=1 \; \text{or} \; \ell=N_L }, \\
\gamma/N_L & & else.
\end{array}
\right.
\end{equation}
where $\gamma$ controls the expected number of unfrozen layers during optimization. Since LISA does not require additional adaptors and only fine-tunes an expected $\gamma$ layers, it notably reduces the memory usage of LLM fine-tuning.

\subsection{Limitations of LISA}
While demonstrating promising results, we observe that the LISA algorithm inherently has two shortcomings that constrain its memory-performance trade-off: 

\begin{table*}[h]
\centering
\caption{Fine-tuning performance of LLaMA2-7B with various dataset. The results are averaged under three random seeds.}
    \vspace{-1em}
\resizebox{0.8\textwidth}{!}{%
\begin{tabular}{@{}lccccccccc@{}}
\toprule
\textbf{Model} & \textbf{Method} & \textbf{BoolQ} & \textbf{PIQA} & \textbf{SIQA} & \textbf{HellaSwag} & \textbf{WinoGrande} & \bf OBQA & \bf Average \\
\midrule
Llama2-7B  & LISA &  82.0 & 79.9 & 33.5 & 59.7
& 79.6 & \bf 38.8 & 62.25\\
Llama2-7B   & LISA-D & \bf 85.1 & \bf 79.9 & \bf 33.8 & \bf 59.8 & \bf 79.7 & 38.4 & \bf 62.78\\
\bottomrule
\end{tabular}%
}
    \vspace{-0.5em}
\label{tab:linear_decreasing}
\end{table*}

\textbf{i. The middle layers of LISA are sampled uniformly, which can result in suboptimal performance.} To verify this point, we conduct a small experiment where we replace the uniform sampling with a very simple baseline, i.e. monotonic decreasing sampling, where the sample probability is monotonically decreasing from shallow layers to deep layers (noted as LISA-D). Table 
\ref{tab:linear_decreasing} shows that this simple sampling method often outperforms uniform sampling, verifying our concern.

\textbf{ii. The sampled layers of LISA are fine-tuned in a full-rank manner, causing a significant memory increase as the number of sampled layers increases.} To illustrate this, we fine-tune LLaMA2-7B on the GSM8K training set and report the GSM8K score and memory usage of LISA with various numbers of sampled layers $\gamma$, as shown in Table \ref{tab:various_gamma}. The memory requirement of LISA rises significantly from 23GB to 36GB as $\gamma$ increases from 1 to 12. Similarly, the performance improves consistently with the increase in sampled layers. Since sampling more layers results in stronger fine-tuning performance, it is crucial to reduce the associated memory overhead as the number of sampled layers grows.

\section{Outlier-weighed Layerwise Sampling (OWS)}
\vspace{-0.5em}
In this section, we introduce our approach, Outlier-weighed Layerwise Low-Rank Projection (\textbf{OWS}). We will discuss the underlying rationales, present preliminary results, and detail the algorithm design.

% \begin{figure}[h]
%     \centering
%     \includegraphics[width=0.8\linewidth]{figs/owl_score.pdf}
%     \caption{OWS Layerwise outlier distribution of LLaMa2 of Equation \ref{eq:lod}. }
%     \label{fig:Enhancer}
% \end{figure}
The above findings shed light on a principle for designing non-uniform layerwise sampling for LLM fine-tuning: layers with higher outlier ratios should be prioritized during the fine-tuning process. This forms the foundation of our proposed method, Outlier-weighed Layerwise Low-Rank Projection (OWS), which we will present in detail.

\textbf{Outlier-Weighed Sampling (OWS).} Although LISA-D achieves good performance, it is more desirable to seek a more principled approach to determine the layerwise sampling probability. In the context of LLMs, we get inspiration from the unique characteristic of LLMs, outliers, defined as
features and weights exhibiting significantly larger magnitudes compared to the majority of others \citep{kovaleva2021bert,puccetti2022outliers,dettmers2022llm}. It has been widely demonstrated that removing outliers significantly degrades the capacity of LLMs \citep{dettmers2022llm}.

\looseness=-1 Our motivation stems from the crucial role outliers play in preserving LLM performance \citep{yin2023outlier}. We hypothesize that layers with more outliers likely contain more essential information, as they have received larger gradients during training. Therefore, we assign higher sampling probabilities to layers with more outliers during fine-tuning, leading to a substantial improvement in performance. To formulate, let us consider the input of a layer as $\mathbf{X}$ with dimensions $(N\times L, C_{\texttt{in}})$, where $N$ and $L$ represent the batch and sequence dimensions, respectively; and the weight matrix $\mathbf{W}$ has dimensions $(C_{\texttt{out}}, C_{\texttt{in}})$. Outlier score of weight $\mathbf{W}_{\texttt{ij}}$ is computed as $ \mbf{A}_{\texttt{ij}} =\|\mbf{X}_{\texttt{j}}\|_{2} \cdot |\mathbf{W}_{\texttt{ij}}|$. Here, $\|\mbf{X}_{\texttt{j}}\|_{2}$ is the $\ell_{2}$ norm of input feature connected to $\mathbf{W}_{\texttt{ij}}$.

\begin{figure*}[h]
    \centering
    \begin{subfigure}[b]{0.45\linewidth}
        \centering
        \includegraphics[width=\linewidth]{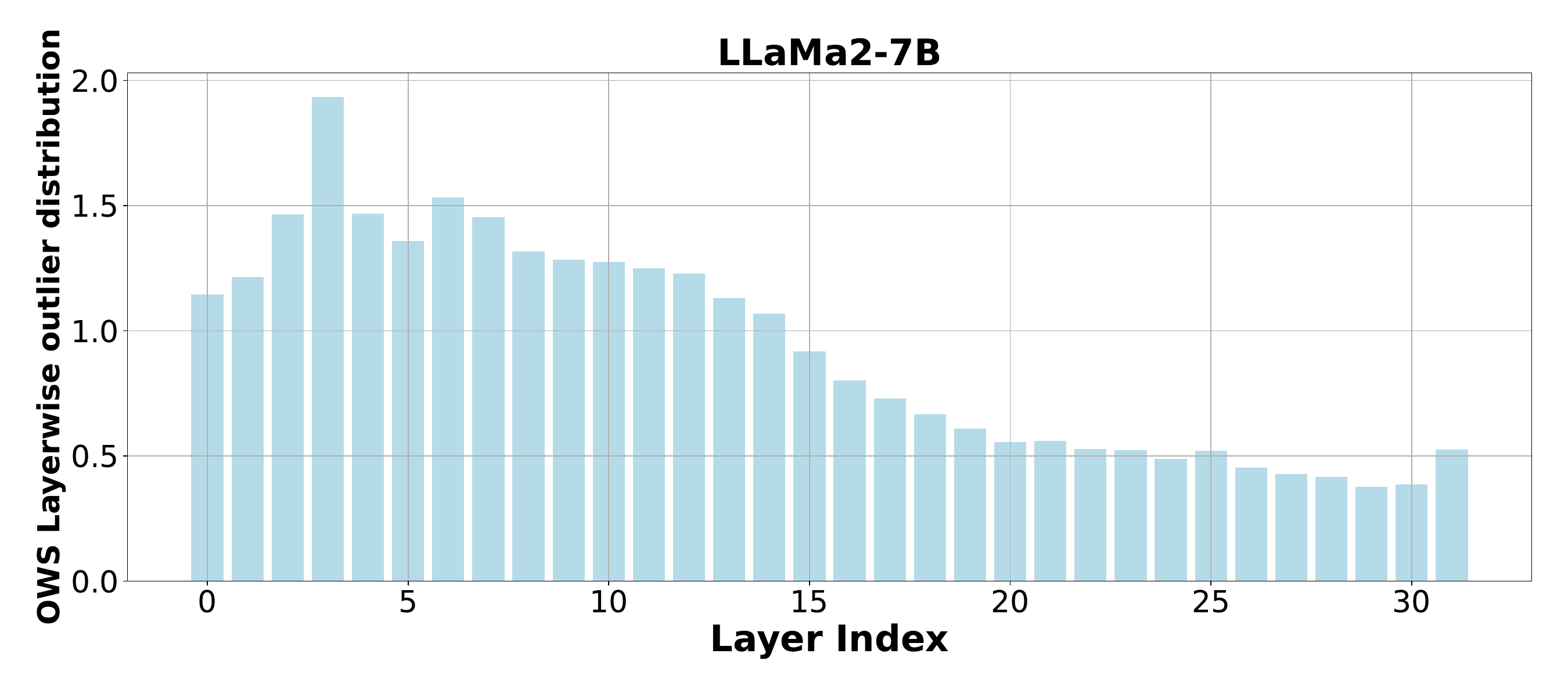}
        % \caption{\textbf{The demonstration of the cost.}}
        \label{fig:cost}
    \end{subfigure}
    \hfill
    \begin{subfigure}[b]{0.45\linewidth}
        \centering
        \includegraphics[width=\linewidth]{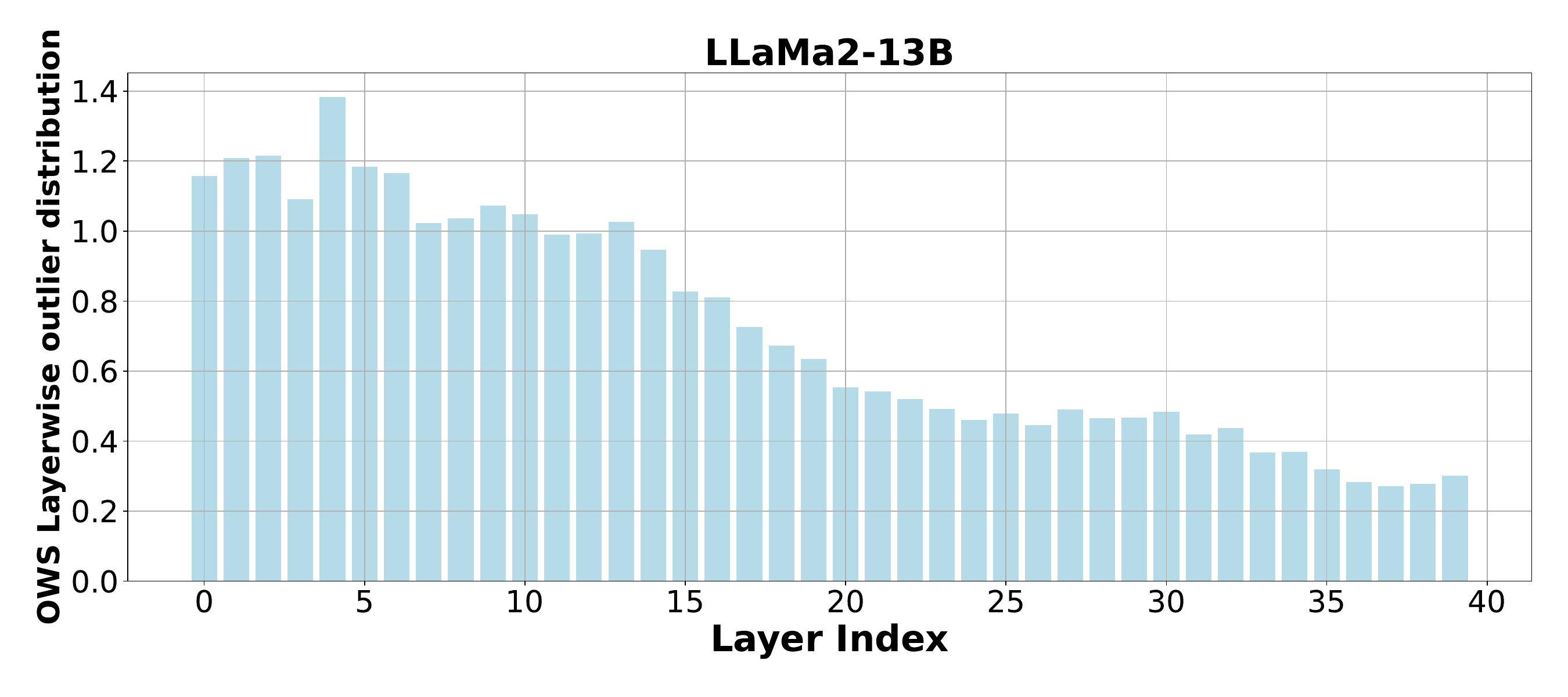}
        % \caption{\textbf{The demonstration of the sample.}}
        \label{fig:sample}
    \end{subfigure}
    \vspace{-1.5em}
    \caption{OWS Layerwise outlier distribution of LLaMa2 of Equation \ref{eq:lod}. The Y-axis is presented in percentage. Higher values mean higher outlier ratios.  }
    \label{fig:LOD}
\end{figure*}

We first calculate the layerwise outlier distribution of 
a $N_L$-layer as $[D_1, D_2, ... , D_{N_L}]$, where $D_{\ell}$ characterizes the outlier ratio of layer $\ell$:
\begin{equation}
    \small 
    D_\ell = \frac{\sum_{i=1}^{C_{\texttt{out}}}\sum_{j=1}^{C_{\texttt{in}}}\mathbb{I}(\mbf{A}^\ell_{\texttt{ij}} > \tau \cdot \mbf{\bar{A}}^\ell)}{C_{\texttt{in}}C_{\texttt{out}}},
    \label{eq:lod}
\end{equation}
where $\mbf{\bar{A}}^\ell$ is the mean of $\mbf{A}^\ell$ and $\mathbb{I}(\cdot)$ is the indicator function, returning 1 if $\mbf{A}^\ell_{\texttt{ij}}$ is larger than $\tau \cdot \mbf{\bar{A}}^\ell$, else 0. The layerwise outlier distribution essentially counts up weights whose outlier score is $\tau$\footnote{We empirically find $\tau=13$ consistently works well and choose it for all experiments in this paper.} times greater than that layer's average outlier score. Larger $D$ means more outliers are presented in the corresponding layer. The sampling probability $p_\ell$ of layer $\ell$ is then calculated as $p_\ell = {\gamma D_\ell/\sum_{i=1}^{N_L}} D_i$, 
where $\gamma$ is the hyperparameter inherited from LISA to control the expected number of unfreeze layers during optimization. At each iteration, only the sampled layers will be fine-tuned, while the remaining layers are kept frozen. The visualization of layerwise outlier distribution of OWS is illustrated in Figure \ref{fig:LOD}.

% OWS naturally leads to a \textit{rich-get-richer}\footnote{\textcolor{blue}{Here, the "rich-get-richer" phenomenon refers to layers with higher initial outlier scores being sampled more frequently for fine-tuning, which leads to these layers being better trained. 

% However, this does not imply that these layers will accumulate more outliers over time as a result of the fine-tuning process.}} phenomenon, where layers containing more outliers during pre-training are sampled and fine-tuned more frequently. 

\textbf{Can the weights in the sampled layer be further updated in a low-rank subspace?} Outlier-weighed sampling addresses our first research question: how to optimally sample layers for sampling-based LLM fine-tuning. To tackle the second issue of the substantial memory cost associated with an increasing number of unfrozen layers, we propose to integrate outlier-weighed sampling with gradient low-rank training. In this approach, the sampled layers are updated in a low-rank manner \cite{zhao2024galore}. Specifically,  for each sampled layer, the gradient matrix is projected into a low-rank subspace using Singular Value Decomposition (SVD). The optimizer states are subsequently updated in the corresponding low-rank subspace with a rank level of $r$, significantly reducing the memory cost of optimization. We update the gradient subspace every 200 iterations to better capture the dynamic trajectory of fine-tuning.  The above two innovations significantly boost the memory efficiency of OWS, unlocking the performance-memory trade-off of sampling-based fine-tuning. At the macro level,  we dynamically sample a limited number of layers to fine-tune at each iteration. At the micro level, each sampled layers are updated with low-rank gradients. 

Since the sampled layers are updated in the low-rank subspace, we can efficiently increase the number of sampled layers $\gamma$ with only a marginal increase in memory cost compared to LISA. Additionally, as we sample only a few layers at each fine-tuning iteration, we can increase the rank levels $r$ without significantly raising the memory requirements compared to LoRA. Memory usage analysis is given in Section 
\ref{sec:memory_usage}. We perform a small search and find that $\gamma=5$ and $r=128$ consistently give us robust performance across models and downstream tasks. Therefore, we choose $\gamma=5$ and $r=128$ as our default settings. We present our
algorithm in Algorithm \ref{alg:OWS}.

%% file: secs/4_exps.tex
% \vspace{-0.2em}
\section{Experiments}
\label{sec:exp}

In this section, we conduct extensive experiments \footnote{Our repository is built on top of LMFlow: https://github.com/OptimalScale/LMFlow} to evaluate the effectiveness of \ours on multiple fine-tuning tasks. Details are provided below.

% \vspace{-0.2em}

\begin{table*}[h]
\centering
\vspace{-0.2em}
\caption{Fine-tuning performance of LLaMa2-7B and Mistral-7B with various approaches on commonsense reasoning datasets. The results are averaged under three random seeds.}
\vspace{-1em}
\resizebox{0.78\textwidth}{!}{%
\begin{tabular}{@{}lcccccccccc@{}}
\toprule
\textbf{Method} & \textbf{Mem.} & \textbf{BoolQ} & \textbf{PIQA} & \textbf{SIQA} & \textbf{HellaSwag} & \textbf{WinoGrande} & \textbf{ARC-e} & \textbf{ARC-c} & \textbf{OBQA} & \textbf{Avg.}\\
\midrule
 \multicolumn{11}{c}{\bf LLaMa2-7B} \\
% \midrule
Full FT & 61G & 87.3 & 79.5 & 32.7 & 56.7 & 80.2 & 78.5 & 49.0 & 40.8 & 63.1  \\
                            LoRA & 26G & 79.7 & 79.7 & 34.4 & 59.9 & 79.8 & 79.5 & 49.7 & 36.6 & 62.4 \\
                            GaLore & 36G & 81.8 & 79.4 & 32.9 & 60.7 & 79.6 & 79.8 & 49.4 & 37.6 & 62.7 \\
                            LISA & 24G & 82.0 & 79.9 & 33.5 & 59.7 & 79.6 & 80.4 & 51.1 & 38.8 & 63.1 \\
                           % & LISA-Decreasing & 24G & 85.1 & 79.9 & 33.8 & 59.8 & 79.7 & 79.8 & 51.0 & 38.4 & 63.4 \\
                           \gr
                           % OWS (Full-Rank)  & 24G & 85.1 & 80.3 & 34.5 & 59.8 & 80.5 & 80.1 & 51.5 & 39.2 & 63.9 \\
                           \gr
                           OWS & \textbf{23G} & 85.4 & 80.7 & 34.2 & 60.3 & 82.2 & 80.6 & 51.0 & 39.1 & \textbf{64.2}\\

\midrule 
 \multicolumn{11}{c}{\bf Mistral-7B} \\
% \midrule
Full FT & 61G & 86.5 & 84.3 & 33.5 & 65.1 & 87.1 & 83.8 & 57.5 & 41.2 & 67.4  \\
                            LoRA & 26G & 87.2 & 81.0 & 33.7 & 62.9 & 83.3 & 82.2 & 54.2 & 37.0 & 65.2 \\
                            GaLore & 36G & 84.8 & 82.5 & 34.4 & 63.5 & 85.6 & 82.5 & 53.9 & 37.8 & 65.6 \\
                            LISA & 24G & 84.7 & 82.9 & 33.4 & 64.2 & 85.8 & 83.4 & 54.4 & 40.5 & 66.2 \\
                             \gr
                            % OWS (Full-Rank) & 24G & 87.3 & 83.8 & 33.7 & 66.1 & 84.9 & 83.7 & 55.3 & 38.2 & 66.7 \\
                             \gr
                            OWS & 23G & 87.8 & 83.9 & 34.0 & 66.4 & 85.6 & 84.1 & 57.9 & 40.4 & \textbf{67.5}\\
\bottomrule
\end{tabular}%
}
\label{table:cs}
\end{table*}

\begin{table*}[htbp]
\centering
\caption{Fine-tuning performance of LLaMa2-7B with various approaches on MT-Bench using GPT-3.5-turbo as a judge. The results are averaged under three random seeds. }
\vspace{-1em}
\resizebox{0.9\textwidth}{!}{%
\begin{tabular}{@{}lccccccccc@{}}
\toprule
 \textbf{Method} & \textbf{Writing} & \textbf{Roleplay} & \textbf{Reasoning} & \textbf{Math} & \textbf{Coding} & \textbf{Extraction} & \textbf{STEM} & \textbf{Humanities} & \textbf{Avg.}\\
\midrule
 Full-FT & 7.11 & 8.11 & 4.90 & 2.85 & 3.75 & 6.50 & 7.80 & 8.10 & 6.14 \\
LoRA &7.21&7.05&4.95&3.25&3.90&5.70&7.90&7.65&5.95\\
GaLore & 7.05 & 7.79 & 3.55 & 2.89 & 3.15 & 6.25 & 8.30 & 7.63 & 5.83 \\
LISA &6.75&7.35&4.35&3.00&3.85&6.85&7.74&7.47&5.92 \\
% & Lisa-Decreasing & 6.85 & 7.58 & 4.85 &3.00 & 4.45 & 6.70 & 7.75 & 7.35 & 6.06\\
\gr
% OWS (Full-Rank) & 7.53 & 8.00 & 4.93 & 3.25 & 4.53 & 6.33 & 8.50 & 8.57 & 6.46\\
\gr
OWS & 8.00 & 7.65 & 4.95 & 3.25 & 4.15 & 7.45 & 8.25 & 8.45 & \textbf{6.52}\\
\bottomrule
\end{tabular}%
}
\vspace{-0.2em}
\label{tab:MT_bench}
\end{table*}

% LlaMa2-7B  & Uniform \citep{pan2024lisa} & 82.0 & 79.9 & 33.5 & 59.7
% & 79.6 &  38.8 & 62.25\\
% LlaMa2-7B   & BI \citep{men2024shortgpt} & 82.8 & 79.6 & 33.2 & 60.3 & 80.4 & 36.6  & 62.15\\
% % Llama2-7B   & Wanda & 82.1 & 79.3 & 33.9 & 59.6 & 79.6 & 37.6 \\
% LlaMa2-7B & RM \citep{samragh2023weight} & 83.4 & 80.4 & 33.1 & 57.7 & 79.8 & 37.4 & 61.97 \\
% \gr
% LlaMa2-7B & OWS (ours) & 85.1 & 80.3 & 34.5 & 59.8 & 80.5 & 39.2 & \bf 63.23 \\
% \bottomrule

\subsection{Experimental Setup}

We choose multiple open-source LLMs that are widely used in research and practice, such as LLaMa2-7B \citep{touvron2023llama} and Mistral-7B \citep{jiang2023mistral}. 

\textbf{Fine-tuning Tasks.} We choose an extensive range of fine-tuning tasks aiming to provide a thorough evaluation of \ours. Our fine-tuning tasks cover three categories: (i) \textbf{Commonsense Reasoning} \citep{hu2023llm}, which includes 8 reasoning tasks including. (ii) \textbf{MT-Bench} \citep{zheng2024judging}, a challenging multi-turn question set to assess the conversational and instruction-following abilities of models. We apply GPT-3.5-turbo and GPT-4o as the judge for MT-Bench; 
(iii) \textbf{MMLU} \citep{hendrycks2020measuring}, a massive multitask test consisting of multiple-choice questions from various branches of knowledge.  We adopt the 5-shot setting for MMLU. For Commonsense Reasoning, all models are first fine-tuned on commonsense170k and then evaluated separately on different tasks, following \citet{hu2023llm}; 
For MT-Bench, we first fine-tune models on the Alpaca GPT-4 dataset \citep{peng2023instruction} and then evaluate on MT-Bench following LISA. The results of MMLU are fine-tuned on the auxiliary training dataset and then evaluated on MMLU with 5 shots. 

\textbf{PEFT Baselines.} We mainly consider four state-of-the-art baselines that are closely related to our approach: (i) \textbf{Full fine-tuning (Full FT)}: all parameters of pre-trained models are fine-tuned. Weights, gradients, and optimization states are maintained with full rank; (ii) \textbf{LoRA}~\cite{hu2021lora}: LoRA introduces additional low-rank adaptors and only fine-tunes adaptors, while maintaining pre-trained weights frozen during training; (iii) \textbf{GaLore} \cite{zhao2024galore}: pre-trained LLMs are fine-tuned with low-rank gradient projection. We follow \cite{zhao2024galore} and set the rank level to 8 for both GaLore and LoRA in all fine-tuning tasks; (iv) \textbf{LISA} \cite{pan2024lisa}: LISA is a sampling-based LLM fine-tuning method, which by default samples 2 layers to fine-tune with full rank at each iteration. GaLore and LISA directly fine-tune pre-trained weights without additional adaptors.

\begin{figure*}[htbp]
    \centering
    \includegraphics[width=.9\linewidth]{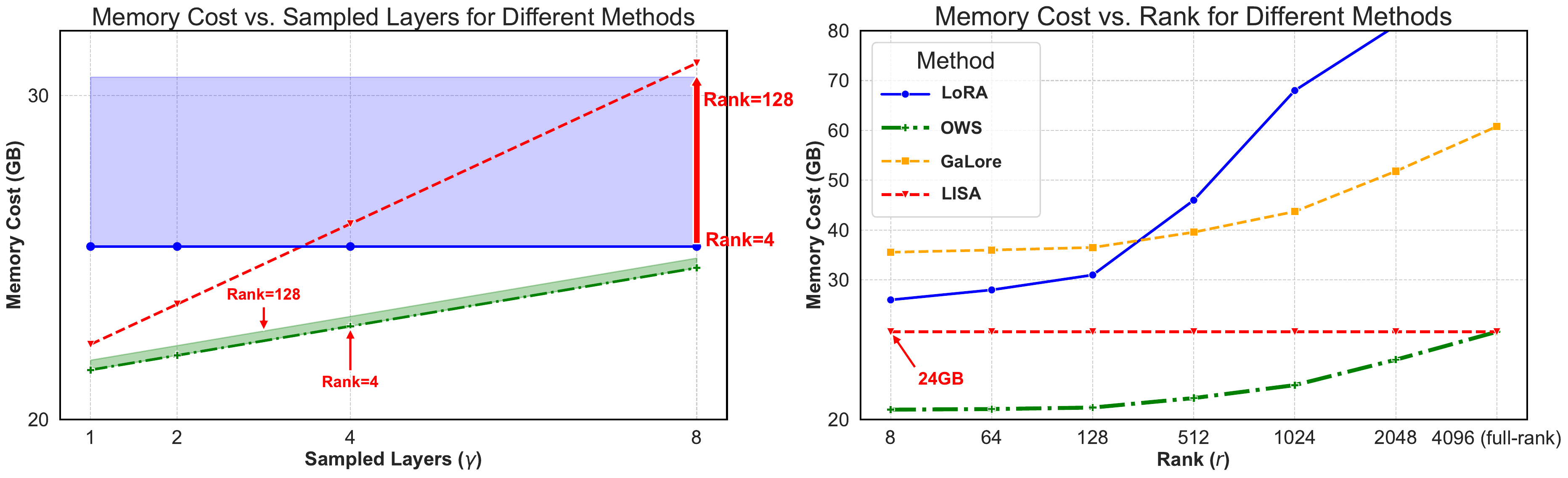}
        \vspace{-1em}
    \caption{Fine-tuning memory usage of using various with LLaMa2-7B. \textbf{Left:} varying sampled layers. In this scenario, we also vary the rank of LoRA and OWS from 4 to 128 to provide a comprehensive analysis. OWS consistently demonstrates superior memory efficiency across all configurations. Notably, LISA's memory advantage over LoRA diminishes as the number of sampled layers increases. \textbf{Right:} varying ranks. The sampled layer of LISA and OWS is set as $\gamma=2$. }
 \vspace{-1em}
    \label{fig:memory}
\end{figure*}

\textbf{Hyperparameter Tuning.} Regarding the hyperparameters of the baselines, we have conducted extensive hyperparameter tuning for all baselines with LLaMa2-7B and reported the results with the best ones. For Mistral-7B, we directly use the best hyperparameters of LLaMa2-7B. Specifically, for the learning rate, we performed a hyperparameter sweep over [1e-4, 3e-4, 7e-5, 5e-5, 1e-5, 5e-6] for each method. For GaLore, we tested several update frequencies for the subspace [50, 100, 200, 500] and found that 200 works best, consistent with GaLore's reports. To ensure a fair comparison, we followed GaLore's approach and set the rank level to 8 for GaLore and LoRA, resulting in approximately 24GB of memory usage for all methods. Additionally, we thoroughly analyzed the effect of two hyperparameters, such as rank level and sampled layers, as shown in Figure \ref{fig:memory}, where our approach consistently demonstrates superior memory benefits. More configurations details are reported in Appendix \ref{app:hyper}.

% \vspace{-1em}

\subsection{Experimental Results}
In this section, we present the empirical results of OWS in comparison to other baseline methods.

\textbf{Commonsense Reasoning Benchmark.} 
We first evaluate with 8 commonsense reasoning tasks. The results are reported in Table \ref{table:cs}. Overall, OWS consistently outperforms Full FT and other PEFT baselines by a large margin across various LLMs, demonstrating the superiority of OWS in LLM fine-tuning. We summarize our key observations below:

\circled{1} \textbf{OWS approaches significantly outperform other efficient fine-tuning approaches by a large margin.} OWS consistently outperforms its layerwise sampling baseline, LISA, on nearly all tasks with LLaMA2-7B, delivering an average of 1.1\% performance gain. 

\circled{2} \textbf{OWS outperforms full fine-tuning across tasks on LLaMa.} We can observe that OWS can achieve better performance than full fine-tuning with all models. LISA can match the performance of full fine-tuning for LLaMa models, whereas GaLore and LoRA perform no better than full fine-tuning. However, only \ours is able to match the performance of full fine-tuning with Mistral-7B and all other baselines fail to do so. This result suggests that LLMs gain greater benefits by leveraging features within important layers rather than uniformly distributing resources across all layers for fine-tuning.

% \circled{3} \textbf{LLaMa3-8B consistently outperforms LLaMa2-7B on Commonsense Reasoning.} LLaMa3-8B consistently outperforms its previous version. Interestingly, performance variance between different fine-tuning approaches of LLaMa3 is smaller than LLaMa2.

% appears to be more effective on some downstream tasks such as BoolQ and WinoGrande

\textbf{MT-Bench.} We next evaluate OWS on a more comprehensive benchmark, MT-Bench, featuring 80 high-quality, multi-turn questions designed to assess LLMs on 8 common categories. Results are presented in Table \ref{tab:MT_bench}. We can observe that the benefits of OWS over other PEFT approaches are more pronounced. Using GPT-3.5-turbo as a judge, all other baselines fail to match the performance of full fine-tuning on MT-Bench with scores below 6.0, whereas OWS outperforms the full fine-tuning by a large margin.  To be specific, OWS  significantly boosts the average score of LISA from 5.92 to 6.52.

\begin{table}[h]
\centering
\caption{Mean score of LLaMA-2-7B on MT-Bench over three seeds. The results are averaged under three random seeds.}
\vspace{-1em}
\resizebox{0.48\textwidth}{!}{%
\begin{tabular}{@{}lccccc@{}}
\toprule
 \textbf{Judge} & \textbf{Full-FT} &  \textbf{LoRA} & \textbf{GaLore} & \textbf{LISA}  & \textbf{OWS} \\
\midrule
 GPT-3.5-turbo & 6.14 & 5.95 & 5.83 & 5.92 &  \textbf{6.52} \\
 GPT-4o & 4.91 & 4.58 & 4.73 & 4.81 &  \bf 5.10  \\

\bottomrule
\end{tabular}%
}
\vspace{-1em}
\label{tab:gpt4}
\end{table}

The performance trend when using GPT-4 is very similar to that of GPT-3.5-turbo, although the scores evaluated by GPT-4 are generally lower. Notably, only OWS outperforms full fine-tuning, achieving a higher score over full fine-tuning.

\textbf{MMLU Benchmark.} To draw a more solid conclusion, we also test another widely used benchmark, i.e., MMLU. The results are shown in Table \ref{tab:MMLU_bench}. Our findings highlight that OWS consistently outperforms Full FT, while other PEFT methods fall short of dense fine-tuning. Specifically, OWS achieves an average score of 52.6, demonstrating significant improvements across various domains such as Humanities, STEM, Social Sciences, and Others. These results underscore OWS's efficacy beyond full fine-tuning while maintaining superior memory efficiency.

\begin{table}[h]
\centering
\caption{Fine-tuning performance of LLaMa2-7B with various approaches on MMLU benchmark. The results are averaged under three random seeds.}
\vspace{-1em}
\resizebox{0.45\textwidth}{!}{
\begin{tabular}{@{}lcccccc@{}}
\toprule
 \textbf{Method} & \textbf{Humanities} & \textbf{STEM} & \textbf{Social.} & \textbf{Other} & \textbf{Avg.}\\
\midrule
    Full-FT & 49.9 & 41.7& 57.5 & 57.0 & 51.5 \\
    LoRA & 46.1 & 40.8 & 56.6 & 56.2 & 49.9 \\
    GaLore & 45.4 & 41.7 & 55.8 & 56.0 & 49.7 \\
    LISA & 44.9 & 41.2 & 54.7 & 57.6 & 49.6 \\
    % \gr OWS (Full-Rank) & 49.1 & 41.3 & 58.8 & 59.1 & 52.1 \\
    \gr OWS & 49.8 & 42.1 & 58.6 & 59.7 & \textbf{52.6}\\
\bottomrule
\end{tabular}%
}
\label{tab:MMLU_bench}
\end{table}

%--------------------------------------------------
\paragraph{GSM8K.}
We extend our evaluation to compare OWS with two recent memory-efficient fine-tuning methods, \textbf{HiFT}~\cite{liu2024hift} and \textbf{MeZO}~\cite{malladi2023fine}, on the GSM8K benchmark.  
Table~\ref{tab:gsm8k_mem_eff} indicates that OWS achieves the best accuracy, surpassing HiFT by \(+1.6\) percentage points (pp) and MeZO by \(+2.5\) pp under the same model size.

\begin{table}[h]
\centering
\caption{GSM8K accuracy (\%) of LLaMA2-7B with different memory-efficient fine-tuning strategies.}
\label{tab:gsm8k_mem_eff}
\setlength{\tabcolsep}{6pt}
\begin{tabular}{lcc}
\toprule
\textbf{Method} & \textbf{Model} & \textbf{GSM8K} \\
\midrule
HiFT (best: RAN) & LLaMA2-7B & 20.3 \\
MeZO             & LLaMA2-7B & 19.4 \\
\rowcolor{gray!15} OWS ($r{=}128,\gamma{=}2$) & LLaMA2-7B & \textbf{21.9} \\
\bottomrule
\end{tabular}
\end{table}

\paragraph{Generalisability to Newer Architectures.}
To examine scalability, we further fine-tune the recent Qwen2.5-7B model on GSM8K and compare OWS with \textbf{DoRA}~\cite{liu2024dora} and \textbf{LISA}.  
As shown in Table~\ref{tab:qwen_gsm8k}, OWS attains the highest score of 83.7 \%, outperforming DoRA by \(+2.5\) pp and LISA by \(+4.0\) pp, indicating strong generalisation to state-of-the-art LLMs.

\begin{table}[h]
\centering
\caption{GSM8K accuracy (\%) on Qwen2.5-7B.}
\label{tab:qwen_gsm8k}
\setlength{\tabcolsep}{6pt}
\begin{tabular}{lcc}
\toprule
\textbf{Method} & \textbf{Model} & \textbf{GSM8K} \\
\midrule
LISA & Qwen2.5-7B & 79.7 \\
DoRA & Qwen2.5-7B & 81.2 \\
\rowcolor{gray!15} OWS  & Qwen2.5-7B & \textbf{83.7} \\
\bottomrule
\end{tabular}
\end{table}
%--------------------------------------------------

% \subsection{Large Scale Fine-Tuning}
\subsection{Memory Efficiency of OWS}
\label{sec:memory_usage}

Thanks to its layerwise sampling and low-rank characteristics, OWS significantly improves the memory efficiency of LLM fine-tuning. To verify, we report the memory cost of various approaches when used to fine-tune LLaMa2-7B, with a token batch size of 1 in Figure \ref{fig:memory}.

On the one hand, the low-rank nature of OWS allows us to unfreeze more layers without a substantial increase in memory cost compared to LISA. As illustrated in Figure \ref{fig:memory}-Left, when increasing $\gamma$ from 1 to 8, LISA exhibits a notable memory growth from 23GB to 32GB, whereas OWS's memory cost slightly increases from 21GB to 25GB. Compared to LoRA with $r=4$, OWS facilitates training with a much higher rank ($r=128$) while still maintaining a lower memory cost.
On the other hand, Figure \ref{fig:memory}-Right demonstrates that OWS enables high-rank training without significantly compromising memory efficiency, in stark contrast to LoRA. It is important to note that we do not utilize the layer-wise weight update technique used in GaLore for the memory measurement, hence the memory cost of GaLore is higher than reported in GaLore.

We further break down the memory usage during LLM fine-tuning, presenting the results in Figure \ref{fig:learning_curve}-Left. For this analysis, the number of fine-tuned layers $\gamma$ is set to 2 for both LISA and OWS, and rank level $r$ is set to 8 for both LoRA and OWS. LoRA incurs a substantial activation memory cost, although its optimizer and gradient memory requirements are relatively small. In contrast, LISA's optimizer memory cost is large because each layer is trained in full rank, yet it benefits from a small activation memory cost. OWS effectively combines the advantages of both methods, inheriting the small activation memory of LISA while significantly reducing the optimizer memory requirement. %Notably, this benefit allows OWS to fine-tune LLaMa2-7B with only 22GB of memory, demonstrating its superior memory efficiency.

\begin{table*}[h]
\centering
\caption{GSM8K scores/memory usage for fine-tuning LLaMA2-7B with various sampled layers $\gamma$. The results are averaged under three random seeds.}
\resizebox{0.7\textwidth}{!}{
\begin{tabular}{lcccccc}
\toprule
\textbf{Method} & \multicolumn{5}{c}{\textbf{Setting and Scores/Memory }} \\ \midrule
\textbf{GaLore} & r=8, $\gamma$=32 & r=16, $\gamma$=32 & r=32, $\gamma$=32 & r=64, $\gamma$=32 & r=128, $\gamma$=32 \\ \midrule
& 19.1/35.6G & 18.8/35.6G & 18.4/35.8G & 18.7/36.0G & 18.2/36.5G \\ \midrule
\textbf{LISA} & r=full, $\gamma$=1 & r=full, $\gamma$=2 & r=full, $\gamma$=4 & r=full, $\gamma$=8 & r=full, $\gamma$=12 \\ \midrule
& 16.8/23G & 18.8/25G & 19.8/27G & 19.9/32G & 21.7/36G \\ \midrule
\textbf{OWS} & r=128, $\gamma$=1 & r=128, $\gamma$=2 & r=128, $\gamma$=4 & r=128, $\gamma$=8 & r=128, $\gamma$=12 \\ \midrule
& 20.0/21G & 21.9/22G & 23.5/23G & 25.7/25G & \textbf{27.8/27G} \\ \bottomrule
\end{tabular}
}
\label{tab:various_gamma}
\end{table*}

\subsection{Superiority of OWS under Varying Hyperparameters Over LISA} 

The primary hyperparameters of LISA, GaLore, and OWS are the number of fine-tuned layers  \(\gamma\), and the rank level within each layer \(r\). To evaluate their effect on the performance of different approaches, we vary these two hyperparameters and report the results in Table \ref{tab:various_gamma}. We set \(\gamma=32\) for GaLore and  \(r=\text{`full rank'}\) for LISA as their default. We see that GaLore's performance does improve as rank levels, having the lowest score across most cases.  Notably, OWS significantly reduces the memory cost compared to LISA alone—reducing from 36G to 27G with \(r=\text{full}, \gamma=12\)—while achieving a significant improvement of 6.1.

\subsection{OWS Serves as A Better Layerwise Important Metric than Others}

OWS serves as a better layer-wise importance metric than previous ones. We compare OWS with other layerwise importance scores for sampling-based fine-tuning, including Uniform \citep{pan2024lisa}, Relative Magnitude (RM) \citep{samragh2023weight} and Block Influence (BI) \citep{men2024shortgpt} in Table \ref{tab:comp_other_score}. OWS consistently performs better than other layer importance scores. Note that reversing OWS gives us the worse performance as shown in Appendix \ref{app:reverse}.

\begin{table}[h]
\centering
\caption{Comparison with other layer-wise
importance metrics, using LLaMA2-7B on Commonsense Reasoning. }
\vspace{-1em}
\resizebox{0.4\textwidth}{!}{
\begin{tabular}{@{}lcc@{}}
\toprule
\textbf{Model} & \textbf{Sampling Method} & \bf Average \\
\midrule
LlaMa2-7B  & Uniform \citep{pan2024lisa} &  62.25\\
LlaMa2-7B   & BI \citep{men2024shortgpt} &  62.15\\
% Llama2-7B   & Wanda & 82.1 & 79.3 & 33.9 & 59.6 & 79.6 & 37.6 \\
LlaMa2-7B & RM \citep{samragh2023weight} & 61.97 \\
\gr
LlaMa2-7B & OWS (ours) & \bf 63.23 \\
\bottomrule
\end{tabular}%
}
\label{tab:comp_other_score}
\vspace{-1em}
\end{table}

\subsection{Memory Usage Breakdown and Training Loss Curve}
\begin{figure*}[h]
    \centering
    % \vspace{-1em}
    \includegraphics[width=.9\linewidth]{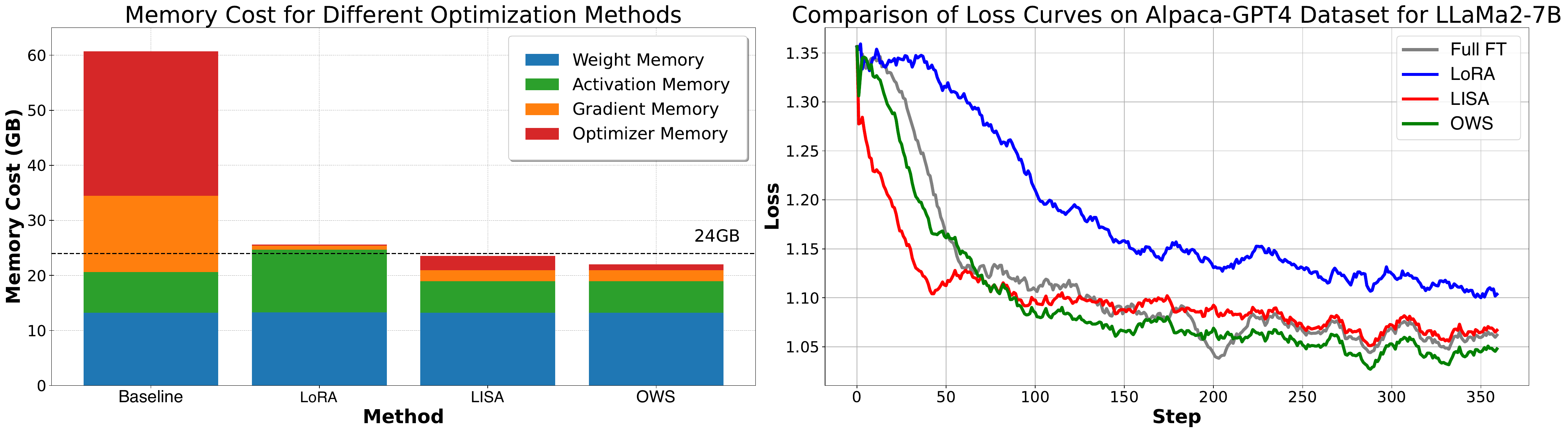}
    % \vspace{-1.6em}
    \caption{\textbf{Left:} Memeory breakdown of various methods using LLaMa2-7B. {\textbf{Right:} Fine-tuning loss of LLaMA2-7B on
     Alpaca GPT-4 dataset using various methods.}}
    \label{fig:learning_curve}
    % \vspace{-0.5em}
\end{figure*}
The training loss curve is an effective way to understand the training dynamics of various methods. Following LISA, we present fine-tuning loss curves of LLaMa2-7B on the Alpaca-GPT4 dataset using Full FT, LoRA, LISA, and OWS in Figure \ref{fig:learning_curve}-Right. At first glance, methods that directly fine-tune pre-trained weights (i.e., LISA and OWS) can better mimic the training landscape of full fine-tuning, compared to LoRA. 

It is worth noting that while OWS initially falls short of LISA in the early phase of training, it gradually catches up after 60 iterations and eventually outperforms LISA with a lower loss. We conjecture that the underlying reason here is that the low-rank update of OWS is less accurate than the full-rank update of LISA at the beginning. However, as training progresses, OWS keeps updating the subspace, leading to an optimal one.

% \vspace{-0.5em}
\section{Related Work}
% \vspace{-0.5em}

\textbf{Parameter-Effieient Fine-Tuning (PEFT).} PEFT is proposed to reduce the prohibitive cost of LLM fine-tuning. Various techniques have been proposed in this dynamic field. For instance, prompt tuning only optimizes input tokens or embeddings while keeping the rest of the model frozen, as demonstrated in studies \citep{lester2021power,li2021prefix,hambardzumyan2021warp,zhong2021factual}. Layer-freezing techniques \citep{liu2021autofreeze,brock2017freezeout,li2024smartfrz} enhance training and fine-tuning efficiency by freezing parts of the layers. Adapter methods \citep{houlsby2019parameter,he2021towards,mahabadi2021parameter,diao2022black}, incorporate a small auxiliary module within the model's architecture, which becomes the exclusive focus of updates during training, thus minimizing the number of trainable parameters and optimizer states.
Among these techniques, Low-Rank Adaptation (LoRA) \citep{hu2021lora} gains massive attention by applying low-rank matrices to approximate weight changes during fine-tuning, which can be merged into the pre-trained weights, leading to no inference overhead. LoRA has been enhanced through various modifications \citep{zhang2023adaptive,renduchintala2023tied,sheng2023s,liu2024dora,kopiczko2023vera,dettmers2024qlora,zhao2024galore} aimed at improving performance and efficiency. Recently, low-rank has also been explored to pre-train LLM from scratch \citep{lialin2023relora,zhao2024galore}.
GaLore \citep{zhao2024galore} projects the gradient into a low-rank subspace for the update to enable full-parameter learning while significantly reducing memory usage during optimization. BAdam \citep{luo2024badam}  partitions the entire model into distinct blocks and utilizes a block coordinate descent framework to update each block individually, either in a deterministic or random sequence.

\looseness=-1 \textbf{Layerwise Sampling for LLM Fine-tuning.} Importance sampling is a powerful statistical technique used in machine learning to estimate properties of a particular distribution by sampling from a different, more convenient distribution. Recently, \citet{pan2024lisa} explored the idea of importance sampling to LLM fine-tuning, with the key idea of sampling only $\gamma$ layers at each step to fine-tuning while keeping the rest of layers frozen. The proposed method, Layerwise Importance Sampled AdamW (LISA), outperforms LoRA by a large margin on various benchmarks and even outperforms full parameters training under certain settings. Inspired by LISA, our paper advances the performance of layerwise sampling for LLM fine-tuning, by addressing a couple of shortfalls of LISA.
\section{Conclusion}

In this paper, we study the sampling-based LLM fine-tuning, where at each iteration, only a few layers are sampled and fine-tuned, instead of the whole model. Specifically, we delve into recently-proposed LISA \citep{pan2024lisa} and unveil two shortcomings that constrain its memory-performance trade-off: (1) The middle layers of LISA are sampled uniformly, which can result in suboptimal performance. (2) The sampled layers of LISA are fine-tuned in a full-rank manner, causing a significant memory increase as the number of sampled layers increases.  To address these challenges, we introduced \textbf{OWS}, which assigns higher sampling probabilities to outlier-rich layers and incorporates low-rank gradient projection for improved memory efficiency. Our experiments on LLaMa2 and Mistral demonstrate that OWS significantly boosts performance while reducing memory usage compared to full-rank fine-tuning.

% To solve these problems, we introduce \textbf{OWS},  a novel fine-tuning method that assigns higher sampling probabilities to these outlier-rich layers. This innovative technique enhances fine-tuning performance while maintaining higher memory efficiency compared to traditional full-rank fine-tuning. The memory efficiency of OWS could be further improved by incorporating Low-Rank gradient projection. Combining sampling-based fine-tuning with gradient low-rank projection not only enhances the performance-memory trade-off of sampling-based fine-tuning but also boosts the effectiveness of gradient low-rank projection in LLM fine-tuning,
% Our experiments across various architectures, including LLaMa2, LLaMa3, and Mistral, demonstrate that OWS achieves significant performance improvements while maintaining higher memory efficiency compared to traditional full-rank fine-tuning. 

% These results highlight OWS's potential to make the deployment of sophisticated language models more practical and accessible, particularly in resource-limited settings. The primary limitation of our work remains the limited exploration of very large-scale LLMs such as those with 70 billion parameters, suggesting an avenue for future research.

%% file: acl_latex.bbl
\begin{thebibliography}{49}
\providecommand{\natexlab}[1]{#1}

\bibitem[{Anil et~al.(2023)Anil, Dai, Firat, Johnson, Lepikhin, Passos, Shakeri, Taropa, Bailey, Chen et~al.}]{anil2023palm}
Rohan Anil, Andrew~M Dai, Orhan Firat, Melvin Johnson, Dmitry Lepikhin, Alexandre Passos, Siamak Shakeri, Emanuel Taropa, Paige Bailey, Zhifeng Chen, et~al. 2023.
\newblock Palm 2 technical report.
\newblock \emph{arXiv preprint arXiv:2305.10403}.

\bibitem[{Biderman et~al.(2024)Biderman, Ortiz, Portes, Paul, Greengard, Jennings, King, Havens, Chiley, Frankle et~al.}]{biderman2024lora}
Dan Biderman, Jose~Gonzalez Ortiz, Jacob Portes, Mansheej Paul, Philip Greengard, Connor Jennings, Daniel King, Sam Havens, Vitaliy Chiley, Jonathan Frankle, et~al. 2024.
\newblock Lora learns less and forgets less.
\newblock \emph{arXiv preprint arXiv:2405.09673}.

\bibitem[{Brock et~al.(2017)Brock, Lim, Ritchie, and Weston}]{brock2017freezeout}
Andrew Brock, Theodore Lim, James~M Ritchie, and Nick Weston. 2017.
\newblock Freezeout: Accelerate training by progressively freezing layers.
\newblock \emph{arXiv preprint arXiv:1706.04983}.

\bibitem[{Brown et~al.(2020)Brown, Mann, Ryder, Subbiah, Kaplan, Dhariwal, Neelakantan, Shyam, Sastry, Askell et~al.}]{brown2020language}
Tom Brown, Benjamin Mann, Nick Ryder, Melanie Subbiah, Jared~D Kaplan, Prafulla Dhariwal, Arvind Neelakantan, Pranav Shyam, Girish Sastry, Amanda Askell, et~al. 2020.
\newblock Language models are few-shot learners.
\newblock \emph{Advances in neural information processing systems}, 33:1877--1901.

\bibitem[{Dettmers et~al.(2022)Dettmers, Lewis, Belkada, and Zettlemoyer}]{dettmers2022llm}
Tim Dettmers, Mike Lewis, Younes Belkada, and Luke Zettlemoyer. 2022.
\newblock Llm. int8 (): 8-bit matrix multiplication for transformers at scale.
\newblock \emph{Advances in Neural Information Processing Systems (NeurIPs)}.

\bibitem[{Dettmers et~al.(2024)Dettmers, Pagnoni, Holtzman, and Zettlemoyer}]{dettmers2024qlora}
Tim Dettmers, Artidoro Pagnoni, Ari Holtzman, and Luke Zettlemoyer. 2024.
\newblock Qlora: Efficient finetuning of quantized llms.
\newblock \emph{Advances in Neural Information Processing Systems}, 36.

\bibitem[{Diao et~al.(2022)Diao, Huang, Xu, Li, Lin, Zhou, and Zhang}]{diao2022black}
Shizhe Diao, Zhichao Huang, Ruijia Xu, Xuechun Li, Yong Lin, Xiao Zhou, and Tong Zhang. 2022.
\newblock Black-box prompt learning for pre-trained language models.
\newblock \emph{arXiv preprint arXiv:2201.08531}.

\bibitem[{Hambardzumyan et~al.(2021)Hambardzumyan, Khachatrian, and May}]{hambardzumyan2021warp}
Karen Hambardzumyan, Hrant Khachatrian, and Jonathan May. 2021.
\newblock Warp: Word-level adversarial reprogramming.
\newblock \emph{arXiv preprint arXiv:2101.00121}.

\bibitem[{He et~al.(2021)He, Zhou, Ma, Berg-Kirkpatrick, and Neubig}]{he2021towards}
Junxian He, Chunting Zhou, Xuezhe Ma, Taylor Berg-Kirkpatrick, and Graham Neubig. 2021.
\newblock Towards a unified view of parameter-efficient transfer learning.
\newblock \emph{arXiv preprint arXiv:2110.04366}.

\bibitem[{Hendrycks et~al.(2020)Hendrycks, Burns, Basart, Zou, Mazeika, Song, and Steinhardt}]{hendrycks2020measuring}
Dan Hendrycks, Collin Burns, Steven Basart, Andy Zou, Mantas Mazeika, Dawn Song, and Jacob Steinhardt. 2020.
\newblock Measuring massive multitask language understanding.
\newblock \emph{arXiv preprint arXiv:2009.03300}.

\bibitem[{Houlsby et~al.(2019)Houlsby, Giurgiu, Jastrzebski, Morrone, De~Laroussilhe, Gesmundo, Attariyan, and Gelly}]{houlsby2019parameter}
Neil Houlsby, Andrei Giurgiu, Stanislaw Jastrzebski, Bruna Morrone, Quentin De~Laroussilhe, Andrea Gesmundo, Mona Attariyan, and Sylvain Gelly. 2019.
\newblock Parameter-efficient transfer learning for nlp.
\newblock In \emph{International conference on machine learning}, pages 2790--2799. PMLR.

\bibitem[{Hu et~al.(2021)Hu, Shen, Wallis, Allen-Zhu, Li, Wang, Wang, and Chen}]{hu2021lora}
Edward~J Hu, Yelong Shen, Phillip Wallis, Zeyuan Allen-Zhu, Yuanzhi Li, Shean Wang, Lu~Wang, and Weizhu Chen. 2021.
\newblock Lora: Low-rank adaptation of large language models.
\newblock \emph{arXiv preprint arXiv:2106.09685}.

\bibitem[{Hu et~al.(2023)Hu, Wang, Lan, Xu, Lim, Bing, Xu, Poria, and Lee}]{hu2023llm}
Zhiqiang Hu, Lei Wang, Yihuai Lan, Wanyu Xu, Ee-Peng Lim, Lidong Bing, Xing Xu, Soujanya Poria, and Roy Ka-Wei Lee. 2023.
\newblock Llm-adapters: An adapter family for parameter-efficient fine-tuning of large language models.
\newblock \emph{arXiv preprint arXiv:2304.01933}.

\bibitem[{Jiang et~al.(2023)Jiang, Sablayrolles, Mensch, Bamford, Chaplot, Casas, Bressand, Lengyel, Lample, Saulnier et~al.}]{jiang2023mistral}
Albert~Q Jiang, Alexandre Sablayrolles, Arthur Mensch, Chris Bamford, Devendra~Singh Chaplot, Diego de~las Casas, Florian Bressand, Gianna Lengyel, Guillaume Lample, Lucile Saulnier, et~al. 2023.
\newblock Mistral 7b.
\newblock \emph{arXiv preprint arXiv:2310.06825}.

\bibitem[{Jiao et~al.(2023)Jiao, Wang, Huang, Wang, Shi, and Tu}]{jiao2023chatgpt}
Wenxiang Jiao, Wenxuan Wang, Jen-tse Huang, Xing Wang, Shuming Shi, and Zhaopeng Tu. 2023.
\newblock Is chatgpt a good translator? yes with gpt-4 as the engine.
\newblock \emph{arXiv preprint arXiv:2301.08745}.

\bibitem[{Kloek and Van~Dijk(1978)}]{kloek1978bayesian}
Teun Kloek and Herman~K Van~Dijk. 1978.
\newblock Bayesian estimates of equation system parameters: an application of integration by monte carlo.
\newblock \emph{Econometrica: Journal of the Econometric Society}, pages 1--19.

\bibitem[{Koco{\'n} et~al.(2023)Koco{\'n}, Cichecki, Kaszyca, Kochanek, Szyd{\l}o, Baran, Bielaniewicz, Gruza, Janz, Kanclerz et~al.}]{kocon2023chatgpt}
Jan Koco{\'n}, Igor Cichecki, Oliwier Kaszyca, Mateusz Kochanek, Dominika Szyd{\l}o, Joanna Baran, Julita Bielaniewicz, Marcin Gruza, Arkadiusz Janz, Kamil Kanclerz, et~al. 2023.
\newblock Chatgpt: Jack of all trades, master of none.
\newblock \emph{Information Fusion}, 99:101861.

\bibitem[{Kopiczko et~al.(2023)Kopiczko, Blankevoort, and Asano}]{kopiczko2023vera}
Dawid~Jan Kopiczko, Tijmen Blankevoort, and Yuki~Markus Asano. 2023.
\newblock Vera: Vector-based random matrix adaptation.
\newblock \emph{arXiv preprint arXiv:2310.11454}.

\bibitem[{Kovaleva et~al.(2021)Kovaleva, Kulshreshtha, Rogers, and Rumshisky}]{kovaleva2021bert}
Olga Kovaleva, Saurabh Kulshreshtha, Anna Rogers, and Anna Rumshisky. 2021.
\newblock Bert busters: Outlier dimensions that disrupt transformers.
\newblock \emph{arXiv preprint arXiv:2105.06990}.

\bibitem[{Lester et~al.(2021)Lester, Al-Rfou, and Constant}]{lester2021power}
Brian Lester, Rami Al-Rfou, and Noah Constant. 2021.
\newblock The power of scale for parameter-efficient prompt tuning.
\newblock \emph{arXiv preprint arXiv:2104.08691}.

\bibitem[{Li et~al.(2024)Li, Yuan, Dai, Zhang, Wang, and Tang}]{li2024smartfrz}
Sheng Li, Geng Yuan, Yue Dai, Youtao Zhang, Yanzhi Wang, and Xulong Tang. 2024.
\newblock Smartfrz: An efficient training framework using attention-based layer freezing.
\newblock \emph{arXiv preprint arXiv:2401.16720}.

\bibitem[{Li and Liang(2021)}]{li2021prefix}
Xiang~Lisa Li and Percy Liang. 2021.
\newblock Prefix-tuning: Optimizing continuous prompts for generation.
\newblock \emph{arXiv preprint arXiv:2101.00190}.

\bibitem[{Lialin et~al.(2023{\natexlab{a}})Lialin, Muckatira, Shivagunde, and Rumshisky}]{lialin2023relora}
Vladislav Lialin, Sherin Muckatira, Namrata Shivagunde, and Anna Rumshisky. 2023{\natexlab{a}}.
\newblock Relora: High-rank training through low-rank updates.
\newblock In \emph{Workshop on Advancing Neural Network Training: Computational Efficiency, Scalability, and Resource Optimization (WANT@ NeurIPS 2023)}.

\bibitem[{Lialin et~al.(2023{\natexlab{b}})Lialin, Shivagunde, Muckatira, and Rumshisky}]{lialin2023stack}
Vladislav Lialin, Namrata Shivagunde, Sherin Muckatira, and Anna Rumshisky. 2023{\natexlab{b}}.
\newblock Stack more layers differently: High-rank training through low-rank updates.
\newblock \emph{arXiv preprint arXiv:2307.05695}.

\bibitem[{Liu et~al.(2024{\natexlab{a}})Liu, Wang, Yin, Molchanov, Wang, Cheng, and Chen}]{liu2024dora}
Shih-Yang Liu, Chien-Yi Wang, Hongxu Yin, Pavlo Molchanov, Yu-Chiang~Frank Wang, Kwang-Ting Cheng, and Min-Hung Chen. 2024{\natexlab{a}}.
\newblock Dora: Weight-decomposed low-rank adaptation.
\newblock In \emph{Forty-first International Conference on Machine Learning}.

\bibitem[{Liu et~al.(2021{\natexlab{a}})Liu, Zheng, Du, Ding, Qian, Yang, and Tang}]{liu2021gpt}
X~Liu, Y~Zheng, Z~Du, M~Ding, Y~Qian, Z~Yang, and J~Tang. 2021{\natexlab{a}}.
\newblock Gpt understands, too. arxiv.
\newblock \emph{arXiv preprint arXiv:2103.10385}.

\bibitem[{Liu et~al.(2024{\natexlab{b}})Liu, Zhang, Li, Liu, Feng, Wang, Zhang, and Sch{\"u}tze}]{liu2024hift}
Yongkang Liu, Yiqun Zhang, Qian Li, Tong Liu, Shi Feng, Daling Wang, Yifei Zhang, and Hinrich Sch{\"u}tze. 2024{\natexlab{b}}.
\newblock Hift: A hierarchical full parameter fine-tuning strategy.
\newblock \emph{arXiv preprint arXiv:2401.15207}.

\bibitem[{Liu et~al.(2021{\natexlab{b}})Liu, Agarwal, and Venkataraman}]{liu2021autofreeze}
Yuhan Liu, Saurabh Agarwal, and Shivaram Venkataraman. 2021{\natexlab{b}}.
\newblock Autofreeze: Automatically freezing model blocks to accelerate fine-tuning.
\newblock \emph{arXiv preprint arXiv:2102.01386}.

\bibitem[{Lu et~al.(2024)Lu, Zhou, Liu, Wang, Mahoney, and Yang}]{lu2024alphapruning}
Haiquan Lu, Yefan Zhou, Shiwei Liu, Zhangyang Wang, Michael~W Mahoney, and Yaoqing Yang. 2024.
\newblock Alphapruning: Using heavy-tailed self regularization theory for improved layer-wise pruning of large language models.
\newblock \emph{Advances in Neural Information Processing Systems}, 37:9117--9152.

\bibitem[{Luo et~al.(2024)Luo, Yu, and Li}]{luo2024badam}
Qijun Luo, Hengxu Yu, and Xiao Li. 2024.
\newblock Badam: A memory efficient full parameter training method for large language models.
\newblock \emph{arXiv preprint arXiv:2404.02827}.

\bibitem[{Mahabadi et~al.(2021)Mahabadi, Ruder, Dehghani, and Henderson}]{mahabadi2021parameter}
Rabeeh~Karimi Mahabadi, Sebastian Ruder, Mostafa Dehghani, and James Henderson. 2021.
\newblock Parameter-efficient multi-task fine-tuning for transformers via shared hypernetworks.
\newblock \emph{arXiv preprint arXiv:2106.04489}.

\bibitem[{Malladi et~al.(2023)Malladi, Gao, Nichani, Damian, Lee, Chen, and Arora}]{malladi2023fine}
Sadhika Malladi, Tianyu Gao, Eshaan Nichani, Alex Damian, Jason~D Lee, Danqi Chen, and Sanjeev Arora. 2023.
\newblock Fine-tuning language models with just forward passes.
\newblock \emph{Advances in Neural Information Processing Systems}, 36:53038--53075.

\bibitem[{Men et~al.(2024)Men, Xu, Zhang, Wang, Lin, Lu, Han, and Chen}]{men2024shortgpt}
Xin Men, Mingyu Xu, Qingyu Zhang, Bingning Wang, Hongyu Lin, Yaojie Lu, Xianpei Han, and Weipeng Chen. 2024.
\newblock Shortgpt: Layers in large language models are more redundant than you expect.
\newblock \emph{arXiv preprint arXiv:2403.03853}.

\bibitem[{Pan et~al.(2024)Pan, Liu, Diao, Pi, Zhang, Han, and Zhang}]{pan2024lisa}
Rui Pan, Xiang Liu, Shizhe Diao, Renjie Pi, Jipeng Zhang, Chi Han, and Tong Zhang. 2024.
\newblock Lisa: Layerwise importance sampling for memory-efficient large language model fine-tuning.
\newblock \emph{arXiv preprint arXiv:2403.17919}.

\bibitem[{Peng et~al.(2023)Peng, Li, He, Galley, and Gao}]{peng2023instruction}
Baolin Peng, Chunyuan Li, Pengcheng He, Michel Galley, and Jianfeng Gao. 2023.
\newblock Instruction tuning with gpt-4.
\newblock \emph{arXiv preprint arXiv:2304.03277}.

\bibitem[{Puccetti et~al.(2022)Puccetti, Rogers, Drozd, and Dell'Orletta}]{puccetti2022outliers}
Giovanni Puccetti, Anna Rogers, Aleksandr Drozd, and Felice Dell'Orletta. 2022.
\newblock Outliers dimensions that disrupt transformers are driven by frequency.
\newblock \emph{arXiv preprint arXiv:2205.11380}.

\bibitem[{Renduchintala et~al.(2023)Renduchintala, Konuk, and Kuchaiev}]{renduchintala2023tied}
Adithya Renduchintala, Tugrul Konuk, and Oleksii Kuchaiev. 2023.
\newblock Tied-lora: Enhacing parameter efficiency of lora with weight tying.
\newblock \emph{arXiv preprint arXiv:2311.09578}.

\bibitem[{Samragh et~al.(2023)Samragh, Farajtabar, Mehta, Vemulapalli, Faghri, Naik, Tuzel, and Rastegari}]{samragh2023weight}
Mohammad Samragh, Mehrdad Farajtabar, Sachin Mehta, Raviteja Vemulapalli, Fartash Faghri, Devang Naik, Oncel Tuzel, and Mohammad Rastegari. 2023.
\newblock Weight subcloning: direct initialization of transformers using larger pretrained ones.
\newblock \emph{arXiv preprint arXiv:2312.09299}.

\bibitem[{Sheng et~al.(2023)Sheng, Cao, Li, Hooper, Lee, Yang, Chou, Zhu, Zheng, Keutzer et~al.}]{sheng2023s}
Ying Sheng, Shiyi Cao, Dacheng Li, Coleman Hooper, Nicholas Lee, Shuo Yang, Christopher Chou, Banghua Zhu, Lianmin Zheng, Kurt Keutzer, et~al. 2023.
\newblock S-lora: Serving thousands of concurrent lora adapters.
\newblock \emph{arXiv preprint arXiv:2311.03285}.

\bibitem[{Surameery and Shakor(2023)}]{surameery2023use}
Nigar M~Shafiq Surameery and Mohammed~Y Shakor. 2023.
\newblock Use chat gpt to solve programming bugs.
\newblock \emph{International Journal of Information technology and Computer Engineering}, (31):17--22.

\bibitem[{Tian et~al.(2023)Tian, Lu, Li, Tang, Cheung, Klein, and Bissyand{\'e}}]{tian2023chatgpt}
Haoye Tian, Weiqi Lu, Tsz~On Li, Xunzhu Tang, Shing-Chi Cheung, Jacques Klein, and Tegawend{\'e}~F Bissyand{\'e}. 2023.
\newblock Is chatgpt the ultimate programming assistant--how far is it?
\newblock \emph{arXiv preprint arXiv:2304.11938}.

\bibitem[{Touvron et~al.(2023)Touvron, Martin, Stone, Albert, Almahairi, Babaei, Bashlykov, Batra, Bhargava, Bhosale et~al.}]{touvron2023llama}
Hugo Touvron, Louis Martin, Kevin Stone, Peter Albert, Amjad Almahairi, Yasmine Babaei, Nikolay Bashlykov, Soumya Batra, Prajjwal Bhargava, Shruti Bhosale, et~al. 2023.
\newblock Llama 2: Open foundation and fine-tuned chat models.
\newblock \emph{arXiv preprint arXiv:2307.09288}.

\bibitem[{Xia et~al.(2024)Xia, Qin, and Hazan}]{xia2024chain}
Wenhan Xia, Chengwei Qin, and Elad Hazan. 2024.
\newblock Chain of lora: Efficient fine-tuning of language models via residual learning.
\newblock \emph{arXiv preprint arXiv:2401.04151}.

\bibitem[{Yin et~al.(2024)Yin, Wu, Zhang, Hsieh, Wang, Jia, Pechenizkiy, Liang, Wang, and Liu}]{yin2023outlier}
Lu~Yin, You Wu, Zhenyu Zhang, Cheng-Yu Hsieh, Yaqing Wang, Yiling Jia, Mykola Pechenizkiy, Yi~Liang, Zhangyang Wang, and Shiwei Liu. 2024.
\newblock Outlier weighed layerwise sparsity (owl): A missing secret sauce for pruning llms to high sparsity.
\newblock \emph{In International Conference on Machine Learning. PMLR.}

\bibitem[{Zhang et~al.(2023)Zhang, Chen, Bukharin, He, Cheng, Chen, and Zhao}]{zhang2023adaptive}
Qingru Zhang, Minshuo Chen, Alexander Bukharin, Pengcheng He, Yu~Cheng, Weizhu Chen, and Tuo Zhao. 2023.
\newblock Adaptive budget allocation for parameter-efficient fine-tuning.
\newblock In \emph{The Eleventh International Conference on Learning Representations}.

\bibitem[{Zhao et~al.(2024)Zhao, Zhang, Chen, Wang, Anandkumar, and Tian}]{zhao2024galore}
Jiawei Zhao, Zhenyu Zhang, Beidi Chen, Zhangyang Wang, Anima Anandkumar, and Yuandong Tian. 2024.
\newblock Galore: Memory-efficient llm training by gradient low-rank projection.
\newblock \emph{arXiv preprint arXiv:2403.03507}.

\bibitem[{Zhao and Zhang(2015)}]{zhao2015stochastic}
Peilin Zhao and Tong Zhang. 2015.
\newblock Stochastic optimization with importance sampling for regularized loss minimization.
\newblock In \emph{international conference on machine learning}, pages 1--9. PMLR.

\bibitem[{Zheng et~al.(2024)Zheng, Chiang, Sheng, Zhuang, Wu, Zhuang, Lin, Li, Li, Xing et~al.}]{zheng2024judging}
Lianmin Zheng, Wei-Lin Chiang, Ying Sheng, Siyuan Zhuang, Zhanghao Wu, Yonghao Zhuang, Zi~Lin, Zhuohan Li, Dacheng Li, Eric Xing, et~al. 2024.
\newblock Judging llm-as-a-judge with mt-bench and chatbot arena.
\newblock \emph{Advances in Neural Information Processing Systems}, 36.

\bibitem[{Zhong et~al.(2021)Zhong, Friedman, and Chen}]{zhong2021factual}
Zexuan Zhong, Dan Friedman, and Danqi Chen. 2021.
\newblock Factual probing is [mask]: Learning vs. learning to recall.
\newblock \emph{arXiv preprint arXiv:2104.05240}.

\end{thebibliography}
